# Do You Trust Me? Cognitive–Affective Signatures of Trustworthiness in Large Language Models

Gerard C. Yeo
National University of Singapore
Singapore
e0545159@u.nus.edu

Svetlana Churina
National University of Singapore
Singapore
churinas@nus.edu.sg

Kokil Jaidka
National University of Singapore
Singapore
jaidka@nus.edu.sg

**Abstract**

Perceived trustworthiness underpins how users navigate online information, yet it remains unclear whether large language models (LLMs),increasingly embedded in search, recommendation, and conversational systems,represent this construct in psychologically coherent ways. We analyze how instruction-tuned LLMs (Llama 3.1 8B, Qwen 2.5 7B, Mistral 7B) encode perceived trustworthiness in web-like narratives using the PEACE-Reviews dataset annotated for cognitive appraisals, emotions, and behavioral intentions. Across models, systematic layer- and head-level activation differences distinguish high- from low-trust texts, revealing that trust cues are implicitly encoded during pretraining. Probing analyses show linearly decodable trust signals and fine-tuning effects that refine rather than restructure these representations. Strongest associations emerge with appraisals of fairness, certainty, and accountability-self– dimensions central to human trust formation online. These findings demonstrate that modern LLMs internalize psychologically grounded trust signals without explicit supervision, offering a representational foundation for designing credible, transparent, and trustworthy AI systems in the web ecosystem. Code and appendix are available at: https://github.com/GerardYeo/TrustworthinessLLM.

**CCS Concepts**

• **Applied computing** → **Psychology**.

**Keywords**

Computational Social Science, Trustworthiness, Large Language Models, Linear Representations, Model Interpretability, Human-AI Alignment

## 1 Introduction

Trustworthiness plays a central role in how users evaluate online information, shaping credibility judgments, sharing decisions, and downstream behavior [8, 13]. As traditional signals of expertise weaken in digital environments [10], these judgments increasingly rely on linguistic cues embedded in user-generated content. At the same time, large language models (LLMs) now mediate search, recommendation, and conversational tasks [7, 17], making it crucial to understand how they internally represent trust-relevant information.

Despite rapid advances in probing and interpretability, we still lack a clear account of whether LLMs encode **perceived trustworthiness** - reflecting *judgments* of sincerity - a concept closely tied to tied to the cognitive and emotional appraisal by the internet user. Prior work shows that models capture sentiment, emotion, and social meaning [2, 4, 5], but no study has examined how trustworthiness itself emerges across model architectures, whether pretraining induces trust-sensitive structure, or how these representations relate to the psychological variables known to guide human trust formation.

To address these gaps, we analyze how modern instruction-tuned LLMs represent perceived trustworthiness in autobiographical online narratives using the PEACE-Reviews dataset [14]. **We focus on three research questions:**[1]

- **RQ1:** How and where do trustworthiness-sensitive representations emerge across transformer depth, and how do these layer–head patterns differ across LLaMA, Qwen, and Mistral architectures?
- **RQ2:** How well trustworthiness can be decoded from model representations? And how fine-tuning on trustworthiness labeled data modify these representational patterns- does it restructure where trust signals reside, or primarily sharpen signals already present in pretrained models?
- **RQ3:** How do internal trustworthiness representations correspond to core psychological constructs–such as fairness, certainty, and accountability–and which constructs are most linearly recoverable from model activations?

## 2 Background

Perceived trustworthiness emerges from a constellation of cognitive, affective, and motivational processes rather than from surface text alone. Research in social cognition shows that readers engage in *third-person inference*—estimating an author's intentions, evaluations, and appraisals—when judging credibility in online environments [6, 12]. Appraisal theory highlights dimensions central to trust formation: *fairness* signals impartiality, *certainty* conveys epistemic stability, and *control* and *accountability* guide interpretations of responsibility and sincerity [3]. Emotional cues such as anger, relief, or joy reveal motivational states and perceived intent, shaping whether authors appear genuine or manipulative [8]. Together, these appraisal and affective signals form the psychological scaffolding through which users interpret credibility in narrative language.

---

[1]Code and appendix are available at: https://github.com/GerardYeo/TrustworthinessLLM.

As large language models (LLMs) increasingly mediate information access, determining whether they encode such psychologically grounded trust cues has become a central interpretability challenge. Much of what humans use to assess trustworthiness manifests not through explicit statements but through discursive and interactional patterns—commitment, distancing, coordination, and conversational asymmetry. Prior work shows that these discourse cues can foreshadow cooperation breakdowns [9] and structure the dynamics of lying and listening in deceptive exchanges [11]. These signals are pragmatic rather than semantic: they arise from how people speak, not just what they say. Although pretrained LLMs capture sentiment, politeness, emotion, and social intent in structured and often linearly decodable subspaces [2, 4, 5], trustworthiness judgments integrate discursive cues with appraisals and inferred behavioral intent in ways that lack a single linguistic signature. It remains unclear whether LLMs internalize these multidimensional trust signals during pretraining, how fine-tuning reshapes them, or how closely model-internal structure aligns with the psychological processes humans use to evaluate credibility.

These questions are especially significant for web-scale applications. Search, recommendation, and conversational systems depend on LLMs both to assess the trustworthiness of user-generated content and to generate outputs that users judge as credible [7, 17]. Yet, progress has been limited by the absence of datasets capturing both linguistic content and the psychological structure underlying trust judgments. The PEACE-Reviews dataset [14, 15] addresses this gap. Its autobiographical narratives contain the appraisals, emotional expressions, and behavioral intentions that readers use to infer credibility, paired with *reader-perceived trustworthiness ratings* and *author-reported appraisals, emotions, and intentions*, all standardized by trained annotators [16]. This dual-perspective structure enables alignment between (a) the psychological mechanisms guiding readers' trust evaluations and (b) the internal states expressed by authors. PEACE-Reviews thus provides an ideal testbed for examining whether trust-sensitive signals emerge in pretrained LLMs, how fine-tuning sharpens them, and how internal representations shape human trust formation.

## 3 Method
### 3.1 Dataset
We use the PEACE-Reviews dataset [14, 15], which contains 1,400 autobiographical narratives describing personal consumption experiences. Each narrative is annotated along two complementary dimensions:

- **Reader-perceived trustworthiness** (5-point Likert)
- **Author-reported psychological variables**, including:
  - Cognitive appraisals (fairness, certainty, control, goal relevance, accountability, normative significance)
  - Emotions (anger, joy, relief, satisfaction, disgust)
  - Behavioral intentions (intention to recommend, purchase intention, perceived helpfulness)

All author-reported labels were coded and verified by trained annotators following the appraisal-theoretic framework in [15, 16]. Narratives are written as free-form autobiographical accounts and retain rich subjective and emotional detail relevant to trustworthiness analysis. All annotated variables are binarized.

### 3.2 Models
We evaluate three state-of-the-art open-weight LLMs: **Llama-3.1-8B-Instruct**, **Mistral-7B-Instruct-v0.3**, and www**Qwen-2.5-7B-Instruct**, to examine how trustworthiness is represented in modern transformer architectures. These models are among the strongest publicly available systems in the 7–8B parameter class, enabling comparison under similar computational scale while avoiding confounds from model-size differences. Using models with comparable capacity allows us to attribute representational differences to architectural and training factors rather than scale, providing a balanced basis for studying whether trustworthiness–and its underlying psychological dimensions–is consistently encoded across contemporary instruction-tuned LLMs.

### 3.3 Experiments
We conduct three categories of analyses aligned with our research questions: (1) whether trustworthiness is represented in intermediate activations, (2) whether such signals are linearly or nonlinearly decodable and how fine-tuning alters them, and (3) whether trust shares representational structure with other psychological constructs.

Narratives are formatted with a consistent chat-style prompt template (Appendix B) and tokenized using each model's tokenizer. Fewer than 2% exceed context limits and are truncated. Activations are extracted via forward passes with no gradient updates. We focus on *pre-projection attention-head outputs*, which capture head-specific computations before mixing through the value projection matrix; prior work shows these activations often correspond to interpretable mechanisms [5].

*Activation Extraction.* For each narrative, we extract the attention-head output for the final token following standard probing practice [1]. This token aggregates contextual information across the sequence and provides a compact summary representation. For each layer $\ell$ and head $h$, we collect the contextualized value-vector outputs used in subsequent analyses.

*RQ1: High–Low Trust Activation Differences.* We assess whether models encode trust-relevant distinctions without supervision by computing the difference in mean activation magnitude between narratives labeled *high* versus *low* trustworthiness (Appendix C). For each layer-head pair, large differences indicate systematic activation shifts that distinguish trust categories, suggesting that latent trust-sensitive features emerge during pretraining.

*RQ2: Trustworthiness Probing.* To evaluate how well trustworthiness can be decoded from intermediate representations, we train logistic regression classifiers for each layer–head pair. Above-chance probe performance indicates that trust information is linearly accessible in the activation space. Probes use an 80/20 train–test split and are evaluated with accuracy and F1 scores.

To examine how supervision modifies these representations, we fine-tune each model using LoRA adapters on trustworthiness labels only and repeat all activation-difference and probing analyses. We compare (a) head-level activation contrasts and (b) layer-level probe accuracy to determine whether fine-tuning restructures trust-related representations or primarily sharpens patterns present in pretrained models.

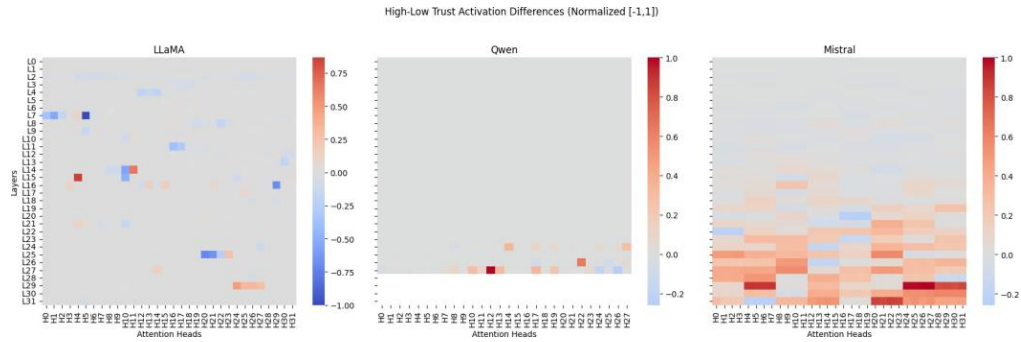

Figure 1: RQ1: Normalized differences in attention-head activations between high- and low-trust narratives across LLaMA, Qwen, and Mistral. Values are normalized within model to the largest absolute difference (range $[-1, 1]$). Positive values indicate stronger activation for high-trust narratives; negative values indicate stronger activation for low-trust narratives.

*RQ3: Probing Psychological Variables.* We extend the probing analysis to all annotated cognitive appraisals, emotions, and behavioral intentions. Because all probes use identical activations, we can directly compare representational strength across constructs and assess how trustworthiness aligns with psychologically grounded dimensions such as fairness, certainty, and accountability.

### 3.4 Robustness Checks

We conduct two robustness analyses to evaluate the stability of trust-related representations across probe types and representational streams:

- **Nonlinear probe formulations**: We train three-layer MLP probes (64-unit hidden layers) on the same activations used for linear probes to test whether trust-relevant information relies on nonlinear separability. Modest improvements over linear probes indicate that trust is encoded largely in linearly accessible subspaces.
- **Post-residual stream analysis**: We probe both the post-attention and post-MLP residual states. Trust remains equally or more decodable from these representations, indicating that trust information is not isolated to attention heads but is integrated into the residual stream and preserved through the block's transformations.

## 4 Results

### 4.1 RQ1: High–Low Trust Activation Differences

Across all three models, we observe clear activation differences between high- and low-trust narratives (Figure 1). This indicates that pretrained LLMs are already sensitive to trust-relevant cues without explicit supervision. The depth at which these signals emerge varies by architecture. Qwen concentrates trust-sensitive activations in its final layers, consistent with late-stage abstraction of evaluative meaning. Mistral shows a mid-to-late pattern, suggesting progressive integration of trust-related cues. LLaMA exhibits the most distributed pattern, with trust signals appearing from early through late layers. This shows that all models encode trust distinctions but allocate them differently across representational depth.

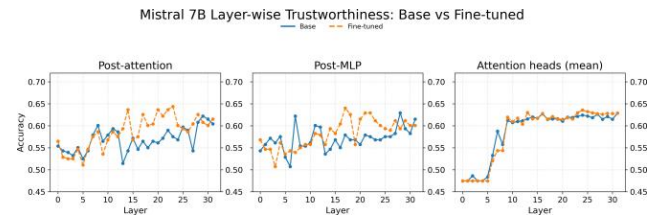

Figure 2: Robustness check: Layer-wise linear probe performance for Mistral 7B: base vs. fine-tuned. Fine-tuning sharpens separability but leaves the representational structure intact.

### 4.2 RQ2: Trustworthiness Probing

We next evaluate whether this structure is linearly recoverable using logistic regression probes trained on each layer–head pair. Probing performance parallels the activation patterns observed in RQ1 (see Appendix D). LLaMA shows broad decodability, peaking at Layer 24, Head 28 (66.5% accuracy; F1 = 66.4). Qwen concentrates its strongest signal in upper layers, with a peak at Layer 25, Head 15 (61.8% accuracy; F1 = 68.8). Mistral shows rising decodability through mid-to-late layers, peaking at Layer 32, Head 9 (66.9% accuracy; F1 = 67.4). These results demonstrate that trustworthiness is linearly accessible across all architectures, with representational depth varying by model design.

Fine-tuning with LoRA improves linear-probe performance at selected layers without shifting where trust signals are concentrated (Figure 2). Head-level activation differences and layer-level probe curves remain structurally similar to the base models. This suggests that supervised fine-tuning sharpens existing trust-relevant distinctions rather than reorganizing representational structure, indicating that models already encode substantial trust-related information during pretraining.

### 4.3 RQ3: Probing Other Psychological Variables

To examine broader psychological structure, we probed all cognitive appraisals, emotions, and behavioral intentions. Figure 3

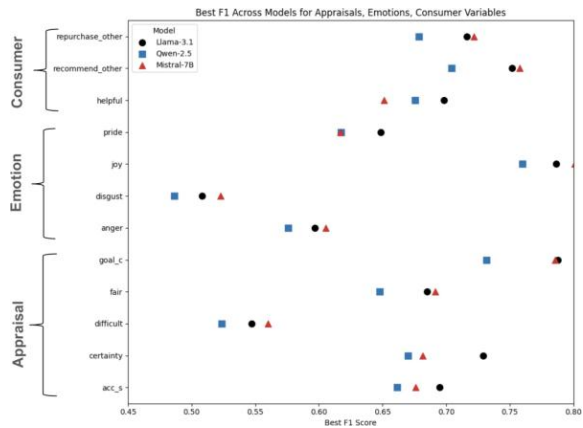

**Figure 3: RQ3: Best F1 score per psychological construct across all layers and heads.** *goal_c, acc_s* refers to goal conduciveness and accountability-self, respectively.

reports the best F1 score achieved across layers and heads for selected constructs (see Appendix H). Constructs central to evaluative judgment–fairness, certainty, accountability–self, goal relevance, and joy–yield strong probe performance across all models. Variables less relevant to trust formation, such as difficulty, control–circumstances, and disgust, show weaker decodability. These patterns align with appraisal theory: the constructs most theoretically tied to trustworthiness are precisely those most robustly encoded by the models.

### 4.4 Robustness Summary

Across all robustness checks–including nonlinear probes and post-residual probing–findings remain stable. Nonlinear probes yield only modest gains over linear probes, indicating that trustworthiness information resides primarily in linearly accessible subspaces. Post-residual analyses confirm that trust-sensitive signals persist after full-layer processing, demonstrating that they are not artifacts of attention-head activations alone. Full robustness results appear in Appendix G.

## 5 Conclusion

This study demonstrates that LLMs encode perceived trustworthiness in systematic, layer- and head-specific patterns. Across LLaMA 3.1, Qwen 2.5, and Mistral 7B, both base and fine-tuned models show consistent representations, with certain layers exhibiting stronger signals that align with cognitive and emotional appraisals. Probing analyses further reveal that trustworthiness is partially linearly accessible but also benefits from nonlinear decoding, indicating that some aspects of trust are embedded in complex representational manifolds. Moreover, trust-related activations show differential relationships with other psychological constructs such as fairness, certainty, and consumer intentions, suggesting that models capture nuanced interactions underlying human judgments.

These findings carry several broader implications. First, they provide a methodological template for interpreting latent representations of socially relevant constructs in LLMs. Second, understanding how trust is encoded can inform the development of AI systems that generate or moderate content responsibly. Finally, linking model representations to psychological theory offers a principled approach for evaluating the alignment of AI behavior with human expectations, highlighting the potential for computational models to support trustworthy, human-centered AI in social and commercial contexts.

## Acknowledgments

This work was supported by the Singapore Ministry of Education AcRF TIER 3 Grant (MOE-MOET32022-0001).

**Table 1: Variables Included for Probing**

| Variables |
|---|
| **Appraisals** |
| Accountability-circumstances |
| Accountability-other |
| Accountability-self |
| Attentional activity |
| Certainty |
| Control-circumstances |
| Control-other |
| Control-self |
| Coping potential |
| Difficulty |
| Effort |
| Expectedness |
| External normative significance |
| Fairness |
| Future expectancy |
| Goal conduciveness |
| Goal relevance |
| Novelty |
| Perceived obstacle |
| Pleasantness |
| **Emotions** |
| Anger |
| Disappointment |
| Disgust |
| Gratitude |
| Joy |
| Pride |
| Regret |
| Surprise |
| **Behavioral Intentions** |
| Intent to promote |
| Intent to repurchase |
| **Consumer-related variables** |
| Helpfulness |
| Trustworthiness |

## A  Probing Variables

Refer to Table 1 for all variables used in the probing experiments. All variables were originally measured on a 5-point scale and subsequently binarized, where responses < 3 were coded as 0 and responses ≥ 3 were coded as 1.

## B  Prompt Templates

To obtain the activations, we parsed the input narratives into the LLMs in the following template. System prompt- "*You are an evaluator trained to assess perceived trustworthiness– that is, whether you think a review seems trustworthy or not*", followed by the user prompt- "*Review: {review text}, After reading the product review, respond with ONLY ONE WORD: 'high' or 'low'. Answer:*".

## C  High–Low Trust Activation Differences Computation

To assess whether perceived trustworthiness is encoded in intermediate representations, we compared mean activation magnitudes for narratives labeled as *high* vs. *low* trustworthiness. For each layer $\ell$ and head $h$, we first extracted the corresponding attention-head output at the final token position. Let $A_{\ell,h}^{(i)} \in \mathbb{R}^d$ denote the activation vector for sample $i$, where $d$ is the head dimensionality.

We then computed the average absolute activation magnitude across samples and token/head dimensions. Formally, we estimate

$$\mu_{\ell,h}^{(\text{high})} = \mathbb{E}_{i \in H}\left[ \|A_{\ell,h}^{(i)}\|_1 \right], \quad (1)$$

and analogously for the low-trust group $\mu_{\ell,h}^{(\text{low})}$. The groupwise activation difference is then:

$$\Delta_{\ell,h} = \mu_{\ell,h}^{(\text{high})} - \mu_{\ell,h}^{(\text{low})}. \quad (2)$$

Thus, $\Delta_{\ell,h}$ measures how strongly attention head ($\ell$, $h$) responds to high- vs. low-trust narratives. Large positive or negative deviations indicate heads whose activation magnitudes distinguish trust categories, suggesting that latent trust-related features emerge even without explicit supervision.

## D  Trustworthiness Probing Results

Figure 4 presents the head-level trustworthiness probing results for LLaMA-3.1, Qwen, and Mistral. Each heatmap visualizes the linear probe accuracy for every attention head across all transformer layers. Brighter (red) regions indicate higher probe accuracy, reflecting heads whose representations contain more trust-relevant information, while darker (blue) regions indicate weaker signal.

Across all three models, trustworthiness is at least partially recoverable from internal representations, though the distribution of probe-sensitive heads varies by architecture. LLaMA-3.1 exhibits a broad, distributed pattern of decodable trust signals across depth. Qwen shows a stronger concentration of high-performing heads in very early and later layers, suggesting a more localized encoding in upper-level representations. Mistral displays a mid-to-late layer progression, with probe accuracy gradually increasing with depth.

These patterns align with the activation-based analyses reported in the main text, indicating consistent architectural differences in where trust-relevant information is stored and how easily it can be linearly extracted.

## E  Comparing base vs. fine-tuned models in trustworthiness probing

Figures 5 and 6 present the comparison between base and fine-tuned models of mistral and qwen in trustworthiness probing.

## F  Fine-Tuning Configuration

We fine-tuned three instruction-tuned large language models: **Qwen2.5-7B-Instruct**, **Mistral-7B-Instruct-v0.3**, and **LLaMA 3.1-8B-Instruct**, using the **Low-Rank Adaptation (LoRA)** technique to enable parameter-efficient fine-tuning. For all models, we set the LoRA rank to $r = 8$, the scaling factor to $\alpha = 32$, and the dropout rate to 0.1, applying adapters to the attention and projection layers (`q_proj`, `k_proj`, `v_proj`, `gate_proj`, `up_proj`, `down_proj`, `o_proj`). Bias terms were disabled (`bias=none`), LoRA weights were initialized randomly, and only the adapter parameters were updated during training while keeping the base model frozen. Training employed a cosine learning rate scheduler with a linear warmup phase, and all experiments were conducted in mixed precision for efficiency. The optimal hyperparameters, including learning rate, batch size, number of epochs, and warmup ratio, were determined through automated search using the **Optuna** framework based on validation loss and model stability across multiple random seeds.

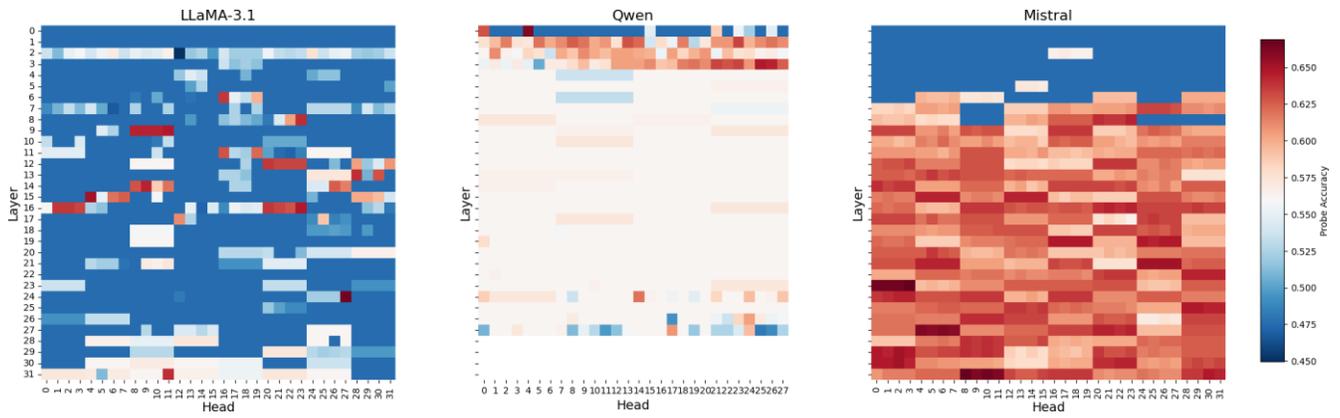

**Figure 4: Head-level trustworthiness probing accuracy for LLaMA-3.1, Qwen, and Mistral. Each heatmap shows probe accuracy (blue = low, red = high) for every attention head across all layers.**

## G  Robustness Analyses

We conducted several robustness checks to evaluate whether our findings about trust representations are consistent across probe types, representational streams, and fine-tuning conditions.

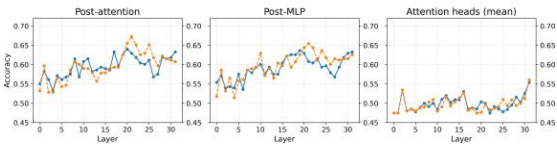

**Figure 5: Line plot comparing the performance of base vs. fine-tuned Llama3.1 models of trustworthiness probes.**

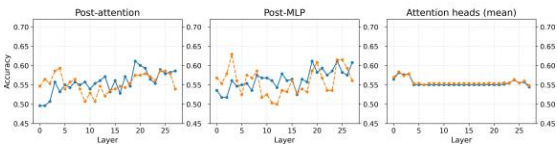

**Figure 6: Line plot comparing the performance of base vs. fine-tuned Qwen 2.5 models of trustworthiness probes.**

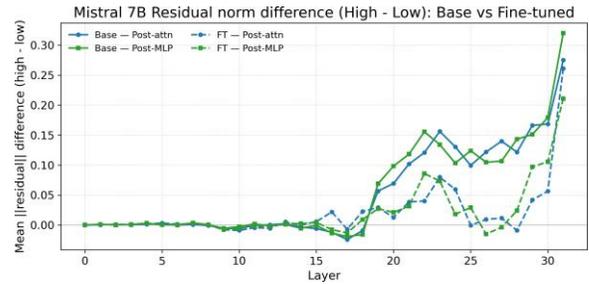

**Figure 8: Mistral**

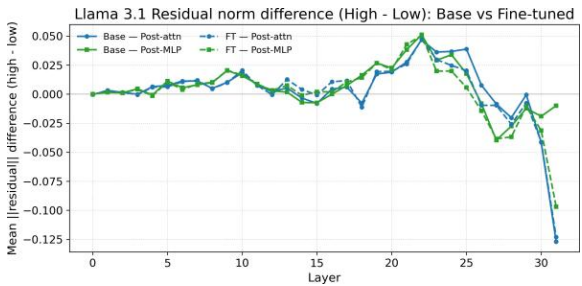

**Figure 7: Llama**

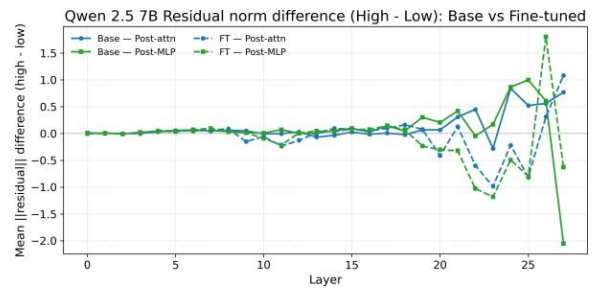

**Figure 9: Qwen**

**Table 2: Performance comparison between base and LoRA fine-tuned models.**

| Model | Base Model | | | Fine-tuned (LoRA) | | |
|---|---|---|---|---|---|---|
| | Acc. | Macro F1 | W-F1 | Acc. | Macro F1 | W-F1 |
| Qwen2.5-7B-Instruct | 0.561 | 0.545 | 0.536 | 0.633 | 0.633 | 0.632 |
| Mistral-7B-Instruct-v0.3 | 0.648 | 0.645 | 0.648 | 0.662 | 0.658 | 0.662 |
| LLaMA 3.1-8B-Instruct | 0.568 | 0.567 | 0.564 | 0.669 | 0.668 | 0.670 |

**Table 3: Per-class F1-scores for the *low* and *high* categories before and after LoRA fine-tuning.**

| Model | Base Model | | Fine-tuned (LoRA) | |
|---|---|---|---|---|
| | F1-Low | F1-High | F1-Low | F1-High |
| Qwen2.5-7B-Instruct | 0.630 | 0.460 | 0.638 | 0.628 |
| Mistral-7B-Instruct-v0.3 | 0.614 | 0.676 | 0.624 | 0.693 |
| LLaMA 3.1-8B-Instruct | 0.595 | 0.539 | 0.652 | 0.685 |

## G.1 Nonlinear Probe Comparison

We evaluated whether nonlinear transformations facilitate trustworthiness decoding. Three-layer MLP probes provided only small gains over linear probes across all models. Peak nonlinear performance for LLaMA (Layer 16, 64.8% accuracy; F1 = 69.0), Qwen (Layer 2, 65.5%; F1 = 67.8), and Mistral (Layer 10, 67.6%; F1 = 69.1) improved only marginally over linear baselines. These results indicate that trust cues occupy largely linearly separable manifolds, with limited benefit from nonlinear transformations.

## G.2 Post-Residual Stream Analysis

Across all three models–Mistral-7B, Qwen-2.5-7B, and LLaMA-3.1 we observe a clear and consistent layerwise pattern: trustworthiness becomes most decodable in the middle-to-late layers of the network, with probe accuracy peaking roughly between Layers 18 and 24. This trend appears regardless of activation type (post-attention or post-MLP), indicating that trust is not encoded primarily in early lexical or syntactic representations. Instead, trust-related features emerge as higher-level semantic abstractions that crystallize in the middle of the transformer stack.

Fine-tuning yields small but uniform accuracy gains across layers, yet the location of the peak remains unchanged, suggesting that fine-tuning sharpens an existing representational structure rather than altering where trust information is formed. This convergence across models and activation streams demonstrates that the middle layers serve as a shared location for trust-related processing in contemporary LLM architectures.

To examine how trust-related information evolves through the model's computation, we measure the difference in residual-stream magnitude between high-trust and low-trust narratives at each layer (Figures 7,8,9). Across all three models, the early layers show near-zero separation, indicating that trust representations are not formed at the lexical or shallow contextual level. Beginning around Layer 15, all models exhibit a clear increase in residual-norm differences, revealing that trust-sensitive features are progressively amplified within the residual stream.

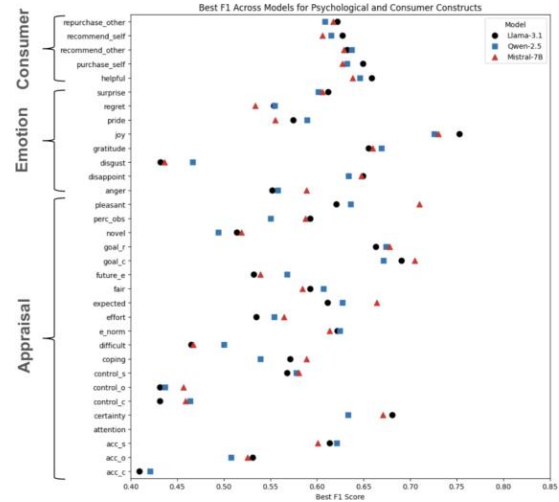

**Figure 10: RQ3: Best F1 score per psychological and consumer-related construct across all layers and heads using post-attention activation states.**

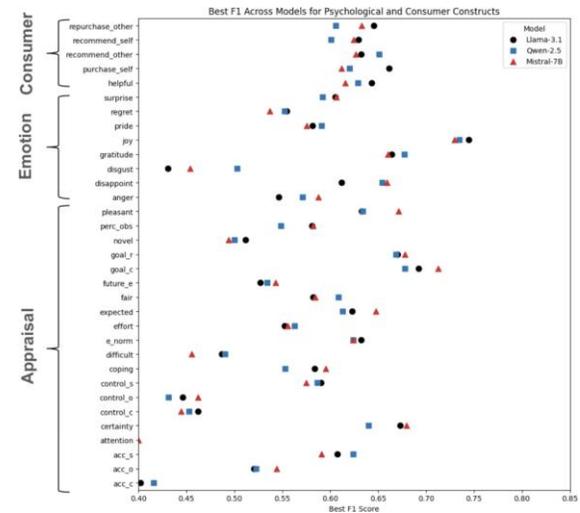

**Figure 11: RQ3: Best F1 score per psychological and consumer-related construct across all layers and heads using post-MLP residual states.**

This amplification peaks in the middle-to-late layers, but with architecture-specific profiles. **Mistral-7B** shows a strong and monotonic increase in residual separation toward the top layers, suggesting that trust becomes increasingly geometrically distinct as the model approaches its output head. **LLaMA-3.1** shows a mid-layer rise followed by a sharp collapse in the final layers, consistent with late-layer compression observed for other semantic features. **Qwen-2.5-7B** shows weak and unstable separation, with small mid-layer differences and large, erratic swings near the top layers, mirroring the weaker trust decoding observed in its probing results.

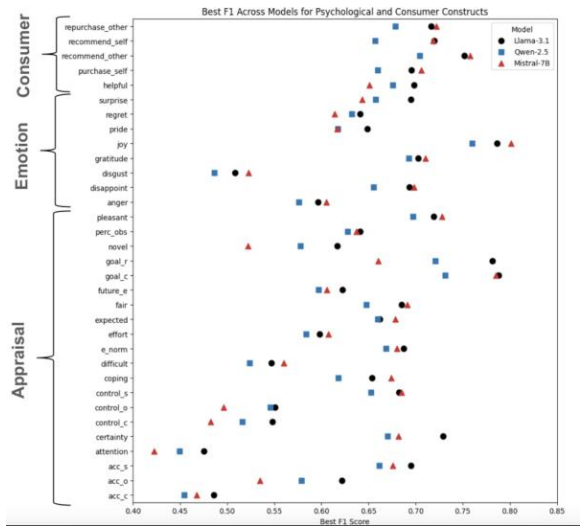

Figure 12: RQ3: Best F1 score per psychological and consumer-related construct across all layers and heads using attention heads.

Fine-tuning slightly elevates the magnitude of high–low differences but preserves each model's characteristic trajectory, indicating that fine-tuning sharpens existing trust-related dynamics rather than altering where they emerge in the computation.

## H   Probing for Psychological and Consumer-related Variables

Figure 12 presents the best F1 scores obtained for each variable across all models when using attention-head activations as features. The results reveal substantial variation in predictability: several variables achieve relatively high F1 scores, indicating strong linear recoverability, whereas others consistently show low scores, suggesting weaker or noisier underlying signals.

Figures 10 and 11 report the corresponding results using post-attention and post-MLP residual streams. These patterns closely mirror those observed in the attention-head analysis, with cognitive appraisals like accountability-self, certainty, goal-related appraisals, and emotions such as joy being best predicted from these trust activations. Taken together, the findings indicate meaningful differences in how well appraisal, emotion, and consumer-related constructs can be predicted from trust-related activations, highlighting that not all constructs are equally linearly recoverable.